# TimeGPT in Load Forecasting: A Large Time Series Model Perspective

Wenlong Liao, Shouxiang Wang, Dechang Yang, Zhe Yang, Jiannong Fang, Christian Rehtanz, Fernando Porté-Agel

***Abstract*—Machine learning models have made significant progress in load forecasting, but their forecast accuracy is limited in cases where historical load data is scarce. Inspired by the outstanding performance of large language models (LLMs) in computer vision and natural language processing, this paper aims to discuss the potential of large time series models in load forecasting with scarce historical data. Specifically, the large time series model is constructed as a time series generative pre-trained transformer (TimeGPT), which is trained on massive and diverse time series datasets consisting of 100 billion data points (e.g., finance, transportation, banking, web traffic, weather, energy, healthcare, etc.). Then, the scarce historical load data is used to fine-tune the TimeGPT, which helps it to adapt to the data distribution and characteristics associated with load forecasting. Simulation results show that TimeGPT outperforms the popular benchmarks for load forecasting on several real datasets with scarce training samples, particularly for short look-ahead times. However, it cannot be guaranteed that TimeGPT is always superior to benchmarks for load forecasting with scarce data, since the performance of TimeGPT may be affected by the distribution differences between the load data and the training data. In practical applications, operators can divide the historical data into a training set and a validation set, and then use the validation set loss to decide whether TimeGPT is the best choice for a specific dataset.**

## I. Introduction

ACCURATE load forecasting is indispensable for the planning and operations of smart grids [1]. For instance, it plays a pivotal role in scheduling generation units, thereby minimizing the need for unnecessary reserve power. Additionally, it enables power system operators to plan the maintenance, ensuring the safe and reliable operation of power systems [2].

Over the past few decades, various approaches to load forecasting have emerged, with traditional methods relying primarily on statistical models. In recent years, however, machine learning models have achieved significant success in various fields [3], [4],[5], driving their application in load forecasting [6].

Specifically, statistical models estimate future load values by analyzing the trends and periodicity in historical data. Classic statistical models include the persistence model (PM), autoregressive moving average, exponential smoothing, autoregressive integrated moving average, linear regression (LR), adaptive filtering model, generalized additive model, and gray model [7]. For example, the work in [8] uses several LR models to forecast the short-term load in California. In [9], the grey model with inverse square root unit functions is introduced to estimate the potential electricity consumption in China over the next few years. In [10], the autoregressive integrated moving average model is designed to forecast long-term loads in Brazil. To mitigate the negative effects of the noise and seasonality in the load data, the work in [11] utilizes exponential smoothing to preprocess the data before conducting load forecasting. These statistical models are grounded in rigorous mathematical principles, offering a high interpretability and relatively low computational costs without requiring extensive historical data. However, their forecast accuracy is limited, especially for a long look-ahead time, due to their difficulty in handling nonlinear relationships [12].

Typically, machine learning models in load forecasting use supervised learning to project the nonlinear relationship between historical load data and forecast values. Popular machine learning models include regression tree (RT), support vector regression, extreme gradient boosting (XGBoost), light gradient boosting machine, multi-layer perceptron (MLP), long short-term memory (LSTM), gated recurrent unit, transformer neural network, convolutional neural network (CNN), and graph neural network [13]. For example, the work in [14] uses the RT model to forecast the short-term load of a city, incorporating the additional information from special days. In [15], the XGBoost model is utilized to forecast peak power demand and long-term electricity consumption, taking into account climatic and economic conditions. In [16], eight meta-heuristic algorithms are adopted to optimize the hyper-parameters of the MLP model, which forecasts the building energy consumption. To capture the temporal features from load data, the LSTM and gated recurrent unit models are presented in [17] and [18], respectively. Simulation analysis shows that LSTM and gated recurrent unit outperform conventional models (e.g., MLP) in short-term load forecasting. Similarly, CNNs and graph neural networks are designed to depict the spatial features from between loads of each bus in

Wenlong Liao, Jiannong Fang, and Fernando Porté-Agel (corresponding author) are with Wind Engineering and Renewable Energy Laboratory, Ecole Polytechnique Federale de Lausanne (EPFL), Lausanne 1015, Switzerland (wenlong.liao@epfl.ch; jiannong.fang@epfl.ch; fernando.porte-agel@epfl.ch).

Shouxiang Wang is with the Key Laboratory of Smart Grid of Ministry of Education, Tianjin University, Tianjin 300072, China (sxwang@tju.edu.cn).

Dechang Yang is with the College of Information and Electrical Engineering, China Agricultural University, Beijing 100083, China (yangdechang@cau.edu.cn).

Zhe Yang is with Department of Electrical and Electronic Engineering, Imperial College London, London SW7 2AZ, United Kingdom (zhe.yang@imperial.ac.uk).

Christian Rehtanz is with the Institute of Energy Systems, Energy Efficiency and Energy Economic, TU Dortmund University, Dortmund 44227, Germany (christian.rehtanz@tu-dortmund.de).

[19], [20]. In general, machine learning models can provide high forecast accuracy, especially when dealing with large-scale load data and long-term forecasts. This is attributed to their ability to automatically extract latent features from data, as well as their good adaptability to complex nonlinear relationships.

However, machine learning models require a large amount of data for training in order to accurately map complex nonlinear relationships. In other words, when historical data is scarce, the forecast accuracy of these machine learning models is limited [21]. For example, in some emerging markets or newly developed communities, inadequate infrastructure may result in a lack of sufficient historical load data. In addition, utilities may not be able to obtain detailed individual electricity consumption data due to privacy concerns [22]. In these scenarios, machine learning models may be constrained by data scarcity, making accurate forecast challenging.

Over the past few years, pre-trained foundation models have significantly driven the rapid development of natural language processing (NLP), computer vision (CV), and speech understanding. For example, large language models (LLMs), such as ChatGPT [23] and Llama [24], perform well in various NLP tasks, even under zero-shot conditions. Similarly, Midjourney [25] and Sora [26] can generate various types of images and videos based on user prompts, respectively. The impressive capabilities of LLMs in CV and NLP have inspired the development and application of foundation models in time series modeling. Recently, foundation models have been extended from CV and NLP to time series analysis. For example, the time series foundation models for transportation and financial problems are presented in [27],[28]. The work in [29] trains a unified time series model to support a universal task specification, including anomaly detection, imputation, classification and prediction. In [30], a time series generative pre-trained transformer (TimeGPT) is presented for time series modeling. Trained on 100 billion data points (e.g., finance, transportation, banking, web traffic, weather, energy, healthcare, etc.), it demonstrated good performance on few-shot learning tasks (e.g., air quality and traffic forecasts). The remarkable success of these foundation models opens up new opportunities for load forecasting, especially in scenarios with scarce historical data.

Inspired by the outstanding performance of LLMs in CV and NLP, this paper aims to discuss the potential of large time series models (LTSMs) in load forecasting with scarce historical data. Specifically, this paper will explore the generalization of TimeGPT to load forecasting, and discuss its advantages and limitations through extensive simulations. The main contributions are as follows:

- **New Perspective:** Unlike classical machine learning models and statistical models characterized by simple structures and few parameters, this paper investigates the potential of LTSMs with complex structures and extensive parameters (i.e., TimeGPT) in load forecasting, from a new perspective.
- **New Application:** By using pre-trained knowledge, the TimeGPT enables load forecasting for scenarios where historical load data is scarce. To our knowledge, this is the first work applying TimeGPT to load forecasting.
- **Extensive Simulations and Practical Suggestions:** Extensive numerical simulations and comprehensive comparisons with eight benchmark models are conducted on five real load datasets to highlight the advantages and limitations of TimeGPT. The actionable suggestions are given for the practical implementation of TimeGPT in load forecasting.

The rest is organized as follows: Section II formulates the TimeGPT. Section III and Section IV conduct simulation and analysis on different datasets. Section V presents the discussion. Finally, Section VI summarizes the conclusion.

## II. Principles and Framework of TimeGPT

This section will formulate the architecture of the TimeGPT, and then present how to train and use it.

### A. Architecture of TimeGPT

Similar to LLMs, the transformer architecture [31] with attention mechanisms is used to construct TimeGPT, as shown in Fig. 1.

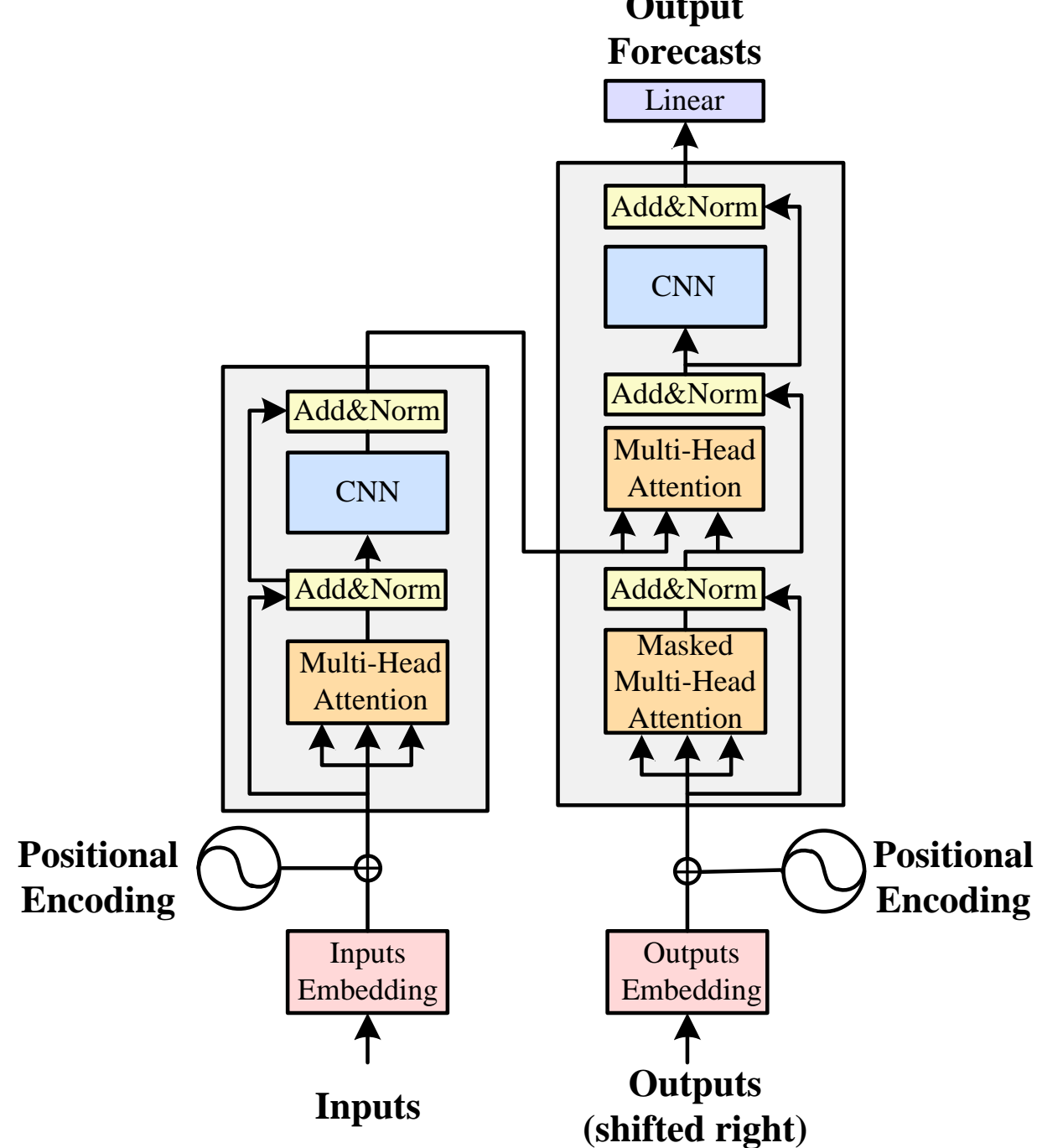

Fig. 1. The basic structure of the transformer block.

In particular, the architecture mainly includes the positional encoding (PE), multi-head attention, and CNN, in which the residual connection and layer normalization are integrated to prevent gradient degradation and accelerate algorithm convergence [30]. By using a segment of historical values, TimeGPT generates forecasts while incorporating local PE for improved input representation. Then, the output of the decoder is projected to forecast values through a linear layer. To maintain this autoregressive property, the input sequence of the decoder is its previously generated tokens, but shifted one position to the right.

Similar to ChatGPT, TimeGPT has the ability to process time series inputs and outputs of varying lengths and frequencies. Firstly, TimeGPT uses the self-attention mechanism of the Transformer model, which enables it to effectively handle time series data of different lengths. The self-attention mechanism dynamically adjusts the focus of TimeGPT based on the global context of the input sequence, ensuring that critical long-term dependencies are not lost when processing long sequences. On the contrary, when dealing with shorter sequences, the self-attention mechanism is still flexible enough to capture short-term dependencies in the sequences. This adaptability allows TimeGPT to effectively handle input data of different lengths, making it suitable for both longer and shorter time series. Secondly, TimeGPT is good at dealing with changes in data frequency. Since the time series data in the pre-training are diverse and may come from different domains with different sampling frequencies, TimeGPT can adapt to changes in frequency when dealing with different time series data. Through fine-tuning, TimeGPT can further understand the characteristics of specific frequency data to make more accurate predictions. In summary, TimeGPT uses the transformer architecture and diverse pre-training data with a strong ability to handle time series data of different lengths and frequencies. This makes it well suited to a variety of time series forecasting tasks, with excellent adaptability and forecasting performance.

***1) Positional Encoding***

To enable the model to accurately understand the sequential information in the input features, the role of the PE is to assign positional information to each feature by using the sine-cosine positional coding [32]:

$$\begin{cases} \mathrm{PE}\left(pos,2i\right)=\sin\left(\dfrac{pos}{10000^{2i/d_{\mathrm{model}}}}\right) \\ \mathrm{PE}\left(pos,2i+1\right)=\cos\left(\dfrac{pos}{10000^{2i/d_{\mathrm{model}}}}\right) \end{cases} \tag{1}$$

where $pos$ denotes the length of the input sequence; $i$ denotes the dimension index of the PE; and $d_{\mathrm{model}}$ denotes the length of the feature.

In PE, $2i$ is controlled by the sine function while $2i+1$ is controlled by the cosine function. The value of the PE changes as the position $pos$ increases. With PE, the model is able to distinguish features at different positions, which leads to a better understanding of the sequential information in the input sequence.

***2) Multi-Head Attention***

As the core unit of the transformer architecture, the multi-head attention can be regarded as an integration of multiple attention heads. As shown in Fig. 2, the ability of the transformer to focus on different features is extended by performing $h$ times attention computations in parallel [33].

In particular, the model learns multiple sets of attention weights simultaneously, and then concatenates their outputs together. Given $h$ attention heads, the calculation of multi-head attention is as follows:

$$\mathrm{MultiHead}\left(Q,K,V\right)=\mathrm{Concat}\left(\mathrm{head}_1,\ldots,\mathrm{head}_h\right)W^O \tag{2}$$

$$\mathrm{head}_i=\mathrm{Attention}\left(Q_i,K_i,V_i\right) \tag{3}$$

$$\mathrm{Attention}\left(Q,K,V\right)=\mathrm{softmax}\left(\frac{QK^T}{\sqrt{d}}\right)V \tag{4}$$

$$\begin{cases} Q=W^Q X \\ K=W^K X \\ V=W^V X \end{cases} \tag{5}$$

where $X$ denotes the input matrix; $W^Q$, $W^K$, and $W^V$ denote the weight matrices of the linear transformation; $Q$, $K$, and $V$ denote the query matrix, key matrix, and value matrix, respectively; and $d$ denotes the dimension of the query matrix.

***3) Convolutional Neural Network***

In the encoder and decoder, a feed-forward neural network (e.g., it is CNN here) is applied to each position to capture the latent feature. The CNN consists of convolutional and pooling layers, which can be formulated as follows [34],[35]:

$$X_{\mathrm{conv,out}}=\sigma\left(W_{\mathrm{conv}}*X_{\mathrm{conv,in}}+B_{\mathrm{conv}}\right) \tag{6}$$

$$X_{\mathrm{pool,out}}=\max_{j,k\in R}\left(X_{\mathrm{pool,in}}^{j,k}\right) \tag{7}$$

where $X_{\mathrm{conv,out}}$ and $X_{\mathrm{pool,out}}$ denote outputs of convolutional and pooling layers, respectively; $X_{\mathrm{conv,in}}$ and $X_{\mathrm{pool,in}}$ denote inputs of convolutional and pooling layers, respectively; σ denotes the activation function; $W_{\mathrm{conv}}$ and $B_{\mathrm{pool}}$ denote weights and bias vectors of the convolutional layer, respectively; $R$ denotes pooling range; and * denotes the convolutional operation.

***4) Residual Connections and Layer Normalization***

To accelerate the convergence of model training, layer normalization (LN) is often used to normalize the outputs of the layers, so that the outputs of each sub-layer remain within a stable range:

$$\mathrm{LN}\left(X_{\mathrm{SL}}\right)=\gamma\frac{X_{\mathrm{SL}}-\bar{X}_{\mathrm{SL}}}{\sqrt{\sigma_X^2+\varepsilon}}+\beta \tag{8}$$

where $X_{\mathrm{SL}}$ denotes the output of the sub-layer; $\bar{X}_{\mathrm{SL}}$ and $\sigma_X$ denote the mean and standard deviation of the output $X_{\mathrm{SL}}$, respectively; $\gamma$ and $\beta$ denote learnable scale and offset parameters, respectively; and $\varepsilon$ denotes a small value to avoid division by zero errors.

To mitigate the problem of vanishing gradients when training deep networks, residual connections are used in each sub-layer:

$$X_{F,out}=F\left(X_{F,in}\right)+X_{F,in} \tag{9}$$

where $X_{\mathrm{F,out}}$ and $X_{\mathrm{F,in}}$ denote the output and input of the sub-layer $F$, respectively.

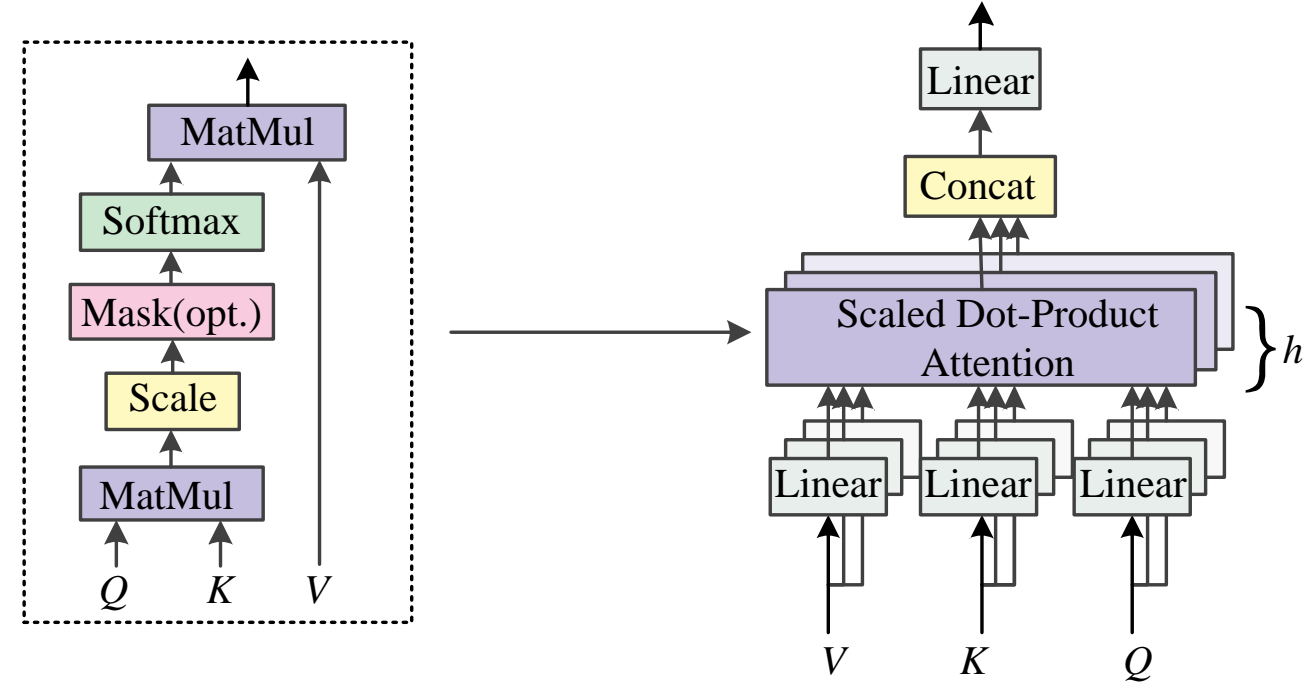

Fig. 2. The basic structure of multi-head attention.

## *B. Bridging Large Models and Load Forecasting*

The large models are designed to handle natural language, which consists of discrete data. In contrast, load forecasting and other time-series forecasting tasks involve time series data, which is inherently continuous. This fundamental difference presents a challenge for large models to directly process time series data.

To bridge this gap, the continuous time series data must be transformed into a discrete format suitable for large models. As shown in Fig. 3(a), this transformation involves two main steps: normalization and quantization.

In the first step, normalization is applied to map the time series data into a specified range to allow faster convergence during model training. Typically, min-max normalization is used, scaling the data between a predefined minimum and maximum, as follows:

$$X' = \left(X - X_{\min}\right) / \left(X_{\max} - X_{\min}\right) \tag{10}$$

where $X_{\min}$ and $X_{\max}$ are the minimum and maximum values, respectively; and $X'$ is the normalized value.

Once the data is normalized, quantization is performed to convert the continuous values into discrete categories. This process uses the widely used equal width binning technique, which segments the normalized data into equal-width intervals. Each interval is assigned a unique value, effectively transforming the continuous values into discrete values compatible with model input, as follows:

$$g\left(X\right) = \begin{cases} 1, & \text{if } X_{\min} \le X < X_{\min} + \Delta d \\ 2, & \text{if } X_{\min} + \Delta d \le x < X_{\min} + 2\Delta d \\ \vdots & \\ m, & \text{if } X_{\min} + (m-1)\Delta d \le x < X_{\max} \end{cases} \tag{11}$$

where $m$ is the number of bin; and $\Delta d=(X_{\max}-X_{\min})/m$ is the width of each bin.

At this point, the continuous time-series data has been fully transformed into a discrete format, allowing it to be processed by large models. Additionally, the output of large models, also in a discrete form, must undergo a reverse transformation to obtain meaningful continuous forecast values, as shown in Fig. 3(b). This reverse process converts the discrete output back into the continuous values required for accurate forecasting.

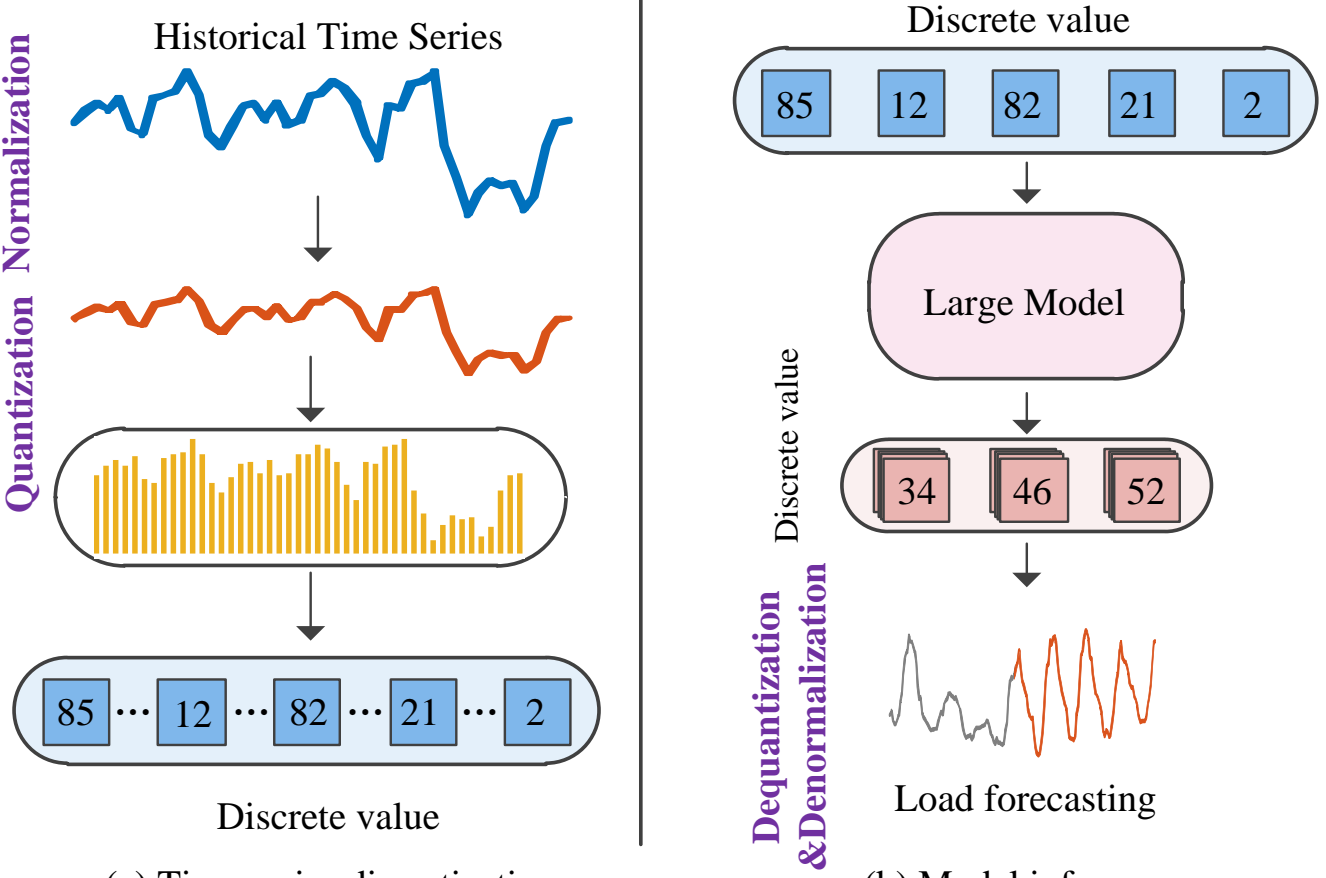

(a) Time series discretization (b) Model inference

Fig. 3. Discretization pipeline for time-series forecasting with large models.

## *C. Training and Use of TimeGPT*

TimeGPT is trained on a large number of publicly available datasets with 100 billion data points, such as finance, transportation, banking, web traffic, weather, energy, healthcare, etc [30]. In regards to temporal characteristics, the training data includes diverse sets with different seasonality, cycles of varying durations, and various trend types. In addition to temporal characteristics, the dataset exhibits variability in noise levels and the presence of outliers, providing a robust training condition. Some datasets exhibit orderly, predictable patterns, while others exhibit significant noise spikes or unexpected events, providing a wide range of scenarios for model assimilation.

TimeGPT has been trained on a cluster of NVIDIA A10G GPUs. The hyper-parameters (e.g. learning rates, batch sizes, etc.) are not open in [30], but they show that a smaller learning rate and a larger batch size are appropriate choices. The deep leaning framework is the PyTorch. The adaptive moment estimation (Adam) is used as the optimizer.

Regarding whether TimeGPT is fine-tuned or not, this paper will consider two scenarios: zero-shot learning and few-shot learning.

For zero-shot learning, the historical load data will be directly fed to TimeGPT to forecast future loads without any adjustments to TimeGPT's parameters.

In the case of few-shot learning, where only scarce historical load data is available due to either inadequate infrastructure or privacy concerns, a fine-tuning process is employed to adapt TimeGPT to the specific load forecasting task. As shown in Fig. 4, the fine-tuning process involves four steps.

Firstly, the pre-trained TimeGPT model is utilized, with all layers having been pre-trained on large and diverse time series datasets. The pre-trained weights in each layer of TimeGPT serve as the starting points for fine-tuning. Secondly, the fine-tuning process is carried out using the scarce historical load data. This data is used to update the weights of all layers in the model. The learning rate for fine-tuning is set lower than typical training processes to ensure that TimeGPT does not deviate significantly from the pre-trained knowledge, but rather adapts to the specific patterns of the new load forecasting task. Thirdly, an appropriate optimization algorithm (e.g., Adam) is used to minimize the loss function, typically mean squared error, over the small dataset. During fine-tuning, TimeGPT is trained on a limited number of epochs to avoid over-fitting given the scarcity of data. Lastly, to prevent over-fitting and ensure generalization, the performance of TimeGPT is monitored on a validation set. Early stopping is applied if the validation performance does not improve after a certain number of epochs, ensuring that the fine-tuned TimeGPT remains robust and does not over-fit the small training dataset.

The fine-tuning ensures that TimeGPT shows good generalization in the given specific task, i.e. load forecasting. Subsequently, the fine-tuned TimeGPT is used to perform load forecasting tasks.

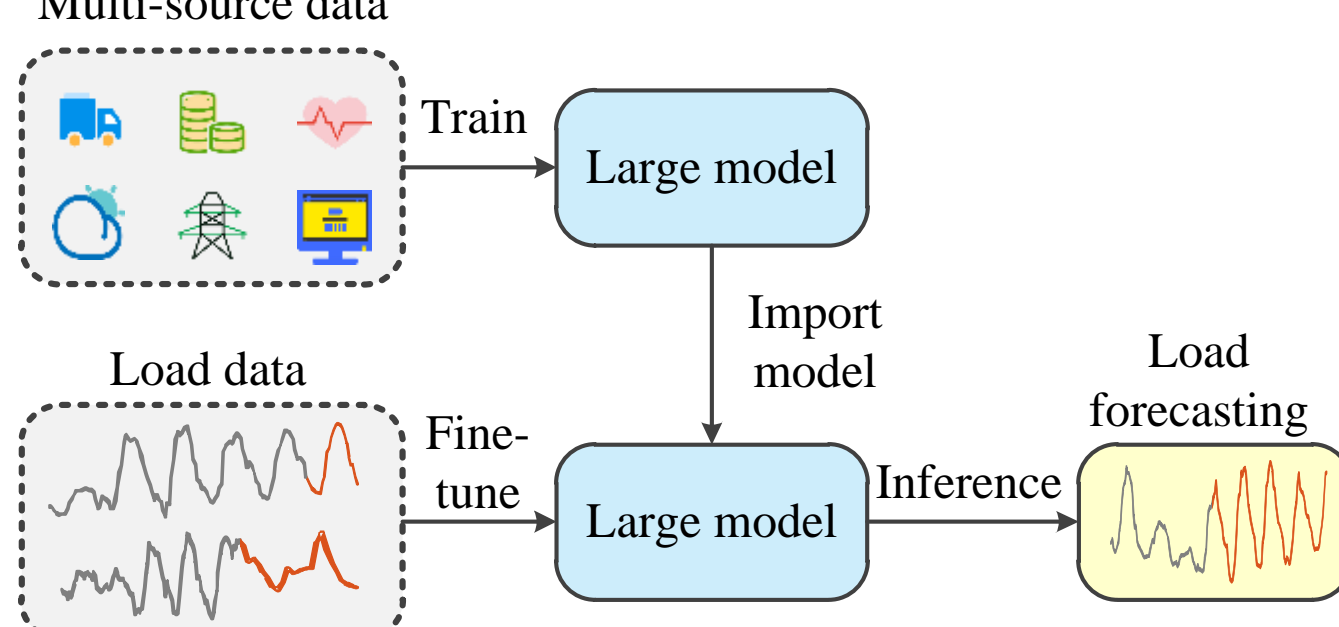

Fig. 4. The fine-tuning process of large models.

## III. Case Study I

This section will conduct simulations and analyses to thoroughly explore the performance of TimeGPT by using a real dataset, while its generalization to other datasets will be tested in the following Section IV.

### A. Simulation Settings

#### 1) Dataset Description

As mentioned earlier, TimeGPT is trained on a large number of publicly available time series datasets. If the popular and publicly available load datasets are used for simulations, there may be a risk of data leakage, since these popular and publicly available datasets may have been used to parameterize TimeGPT. Therefore, simulations are conducted on a private dataset sourced from the University of Texas at Austin [36], ensuring fairness as the parameters of TimeGPT are unrelated to this dataset.

Specifically, this dataset records the load data of 16 campus buildings with a time resolution of one hour. Due to limited application programming interface (API) tokens for TimeGPT, three months of load data are used for simulation and testing, spanning from July 17, 2011, to October 16, 2011.

To test the model performance of TimeGPT on datasets with different numbers of training samples, the original dataset is divided into five cases, as shown in Table I. From case 1 to case 5, the number of training samples gradually increases. Case 1 contains very few training samples, while the training samples in case 5 are relatively rich.

TABLE I
DESCRIPTION OF EACH CASE

| Case | Training set | Test set |
|---|---|---|
| Case 1 | From Jul. 17 to Jul. 19 (3 days) | From Jul. 20 to Oct. 16 |
| Case 2 | From Jul. 17 to Jul. 21 (5 days) | From Jul. 22 to Oct. 16 |
| Case 3 | From Jul. 17 to Jul. 23 (7 days) | From Jul. 24 to Oct. 16 |
| Case 4 | From Jul. 17 to Jul. 31 (15 days) | From Aug. 1 to Oct. 16 |
| Case 5 | From Jul. 17 to Aug. 15 (30 days) | From Aug. 16 to Oct. 16 |

#### 2) Benchmarks

To fully evaluate the performance of TimeGPT, it will be compared to popular models, including PM, LR in [8], RT in [14], XGBoost in [15], MLP in [16], LSTM in [17], patch time series transformer (PatchTST) [37], and time series large language model (TimeLLM) [38]. Note that TimeLLM is trained on massive textual data, while TimeGPT is directly trained on time series data. This key difference means that while TimeLLM is reprogrammed to analyze time series data as textual input, TimeGPT is inherently designed to process time series information from the start, optimizing its parameters specifically for time-dependent data.

Although each model contains numerous hyper-parameters, the impact of these hyper-parameters on performance is not investigated for two main reasons. Firstly, space does not permit a detailed discussion of these hyper-parameters. Secondly, the discussion of these hyper-parameters has already been thoroughly covered in previous works (as cited). The focus of this paper is on the performance of TimeGPT, not on the fine-tuning of the baseline models. To ensure a fair comparison, the Bayesian optimization in [39] is used to determine the most appropriate parameter settings for each baseline model. For example, the parameters of each model for load forecasting with a 1-hour look-ahead time in Case 1 are shown in Table II. For each model, the inputs include only historical loads and time (e.g., data points in the last 24 hours), while the outputs are future loads. It is univariate time series forecasts without considering other information, such as weather conditions. The parameters in other cases can be determined similarly.

TABLE II
PARAMETER OF BENCHMARKS

| Model | Model structure | Fitting setting |
|---|---|---|
| MLP in [16] | Dense 1: 16 units<br>Dense 2: 16 units<br>Dense 3: 1 unit | Training epoch: 200<br>Batch size: 8<br>Optimizer: Adam<br>Learning rate: 0.001<br>Activation function: ReLU for input and middle layers,<br>Sigmoid for the last layer |
| LSTM in [17] | LSTM 1: 16 units<br>LSTM 2: 8 units<br>Dense 1: 8 units<br>Dense 2: 1 unit | |
| PM | Without parameters; It forecast the load by copying the value from the previous time step. | |
| LR in [8] | Intercept is used in calculations | |
| RT in [14] | Max depth: 4<br>Leaves: 25<br>Learning rate:0.01<br>Estimators: 500<br>Min child samples: 90<br>Subsample: 0.8<br>Early stopping: 400<br>Loss: squared error | |
| XGBoost in [15] | | |
| PatchTST in [37] | Length of the patch: 16<br>Stride of the patch:8<br>Number of time units: 96<br>Size of hidden layers in the Transformer: 32<br>Number of attention heads: 4<br>Number of epochs: 200<br>Batch size: 8 | |
| TimeLLM in [38] | Length of the patch: 16<br>Stride of the patch:8<br>Top tokens to consider:5<br>Hidden dimension of LLM: 768<br>Number of heads in attention layer: 8<br>Maximum number of training steps:1000<br>Encoder input size:7<br>Batch size: 8 | |

#### 3) Evaluation Metrics

To avoid chance or coincidence, each model is run 30 times to obtain average forecasts. Then, the model performance is evaluated by using widely used metrics, including mean absolute error (MAE), root mean squared error (RMSE), and mean absolute percentage error (MAPE):

$$\text{MAE} = \frac{1}{n}\sum_{i=1}^{n}\left|y_i - \hat{y}_i\right| \tag{12}$$

$$\mathrm{RMSE}=\sqrt{\frac{1}{n}\sum_{i=1}^{n}\left(y_i-\hat{y}_i\right)^2} \tag{13}$$

$$\mathrm{MAPE}=\frac{1}{n}\sum_{i=1}^{n}\left|\frac{y_i-\hat{y}_i}{y_i}\right| \tag{14}$$

where $y_i$ and $\hat{y}_i$ denotes the normalized real and forecast values, respectively; and $n$ denotes the number of data points in the test set.

***4) Simulation Designs***

To investigate the potential of TimeGPT in load forecasting, simulations will be carried out from the following three points of view:

Firstly, the model performance of TimeGPT will be discussed with and without fine-tuning in load forecasting, i.e., zero-shot learning and few-shot learning. Secondly, TimeGPT will be compared to the benchmarks in cases where historical data is scarce. Finally, TimeGPT will also be compared to the benchmarks in data-rich cases.

## *B. Performance Analysis With and Without Fine-Tuning*

Zero-shot learning involves directly feeding historical load data directly into TimeGPT to forecast future loads without adjusting TimeGPT's parameters. Conversely, few-shot learning involves using scarce historical load data to fine-tune TimeGPT's weights before conducting load forecasting.

To evaluate the model performance of TimeGPT in load forecasting, both few-shot learning and zero-shot learning (i.e., TimeGPT with and without fine-tuning) are considered in five cases. These cases involve load forecasting with different look-ahead times ranging from 1 hour to 24 hours. The average evaluation metrics of TimeGPT in various scenarios are shown in Fig. 5.

Although TimeGPT is trained on massive and diverse datasets consisting of 100 billion data points, its performance on various metrics (e.g., RMSE, MAE, and MAPE) is poor prior to fine-tuning, indicating that it cannot be directly generalized to load forecasting. This could be attributed to the fact that the training data may not adequately represent the specific data distribution and patterns relevant to load forecasting, resulting in limited generalization capability.

However, after fine-tuning, the performance of TimeGPT in load forecasting improves significantly, with considerable decreases observed in various metrics. The reason for this is that the fine-tuning involves specific adjustments to its weights tailored to the requirements of load forecasting, allowing it to better adapt to the data distribution, patterns, and characteristics associated with load forecasting. Therefore, the fine-tuning can significantly improve the performance of TimeGPT, making it more practical and accurate for real-world applications.

## *C. Performance Comparison in Data Scarce Cases*

To compare the performance of TimeGPT with the benchmarks in cases where data is scarce, these models are used to conduct the load forecasting in Cases 1-3 (Data points in the training set range from 3 to 7 days). Note that TimeGPT is fine-tuned by the historical load data here. The average evaluation metrics of TimeGPT in various scenarios are shown in Tables III-V.

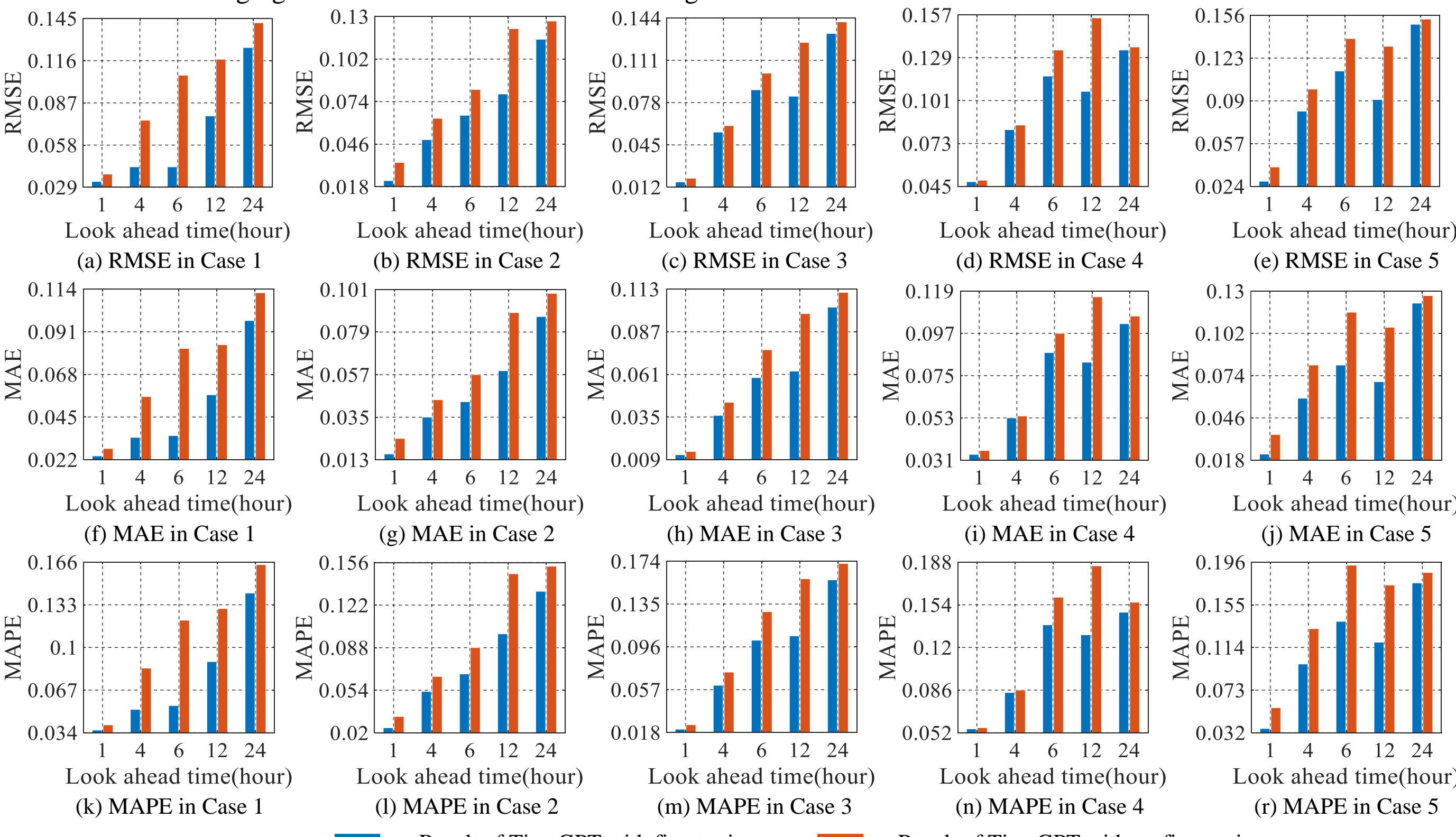

(a) RMSE in Case 1 (b) RMSE in Case 2 (c) RMSE in Case 3 (d) RMSE in Case 4 (e) RMSE in Case 5

(f) MAE in Case 1 (g) MAE in Case 2 (h) MAE in Case 3 (i) MAE in Case 4 (j) MAE in Case 5

(k) MAPE in Case 1 (l) MAPE in Case 2 (m) MAPE in Case 3 (n) MAPE in Case 4 (r) MAPE in Case 5

Fig. 5. The results of TimeGPT with and without fine-tuning in different cases.

TABLE III

THE RESULTS OF MACHINE LEARNING MODELS IN LOAD FORECASTING WITH A SHORT LOOK-AHEAD TIME (CASE 1)

| Model | Look-ahead time=1 hour | | | Look-ahead time=4 hours | | | Look-ahead time=6 hours | | | Look-ahead time=12 hours | | | Look-ahead time=24 hours | | |
|---|---|---|---|---|---|---|---|---|---|---|---|---|---|---|---|
| | RMSE | MAE | MAPE | RMSE | MAE | MAPE | RMSE | MAE | MAPE | RMSE | MAE | MAPE | RMSE | MAE | MAPE |
| TimeGPT | **0.033** | **0.024** | **0.036** | **0.043** | **0.034** | **0.052** | **0.043** | **0.035** | **0.055** | **0.072** | **0.055** | **0.088** | 0.125 | 0.097 | 0.142 |
| MLP | 0.051 | 0.046 | 0.075 | 0.070 | 0.057 | 0.091 | 0.089 | 0.063 | 0.094 | 0.099 | 0.077 | 0.123 | 0.091 | 0.073 | 0.119 |
| LSTM | 0.043 | 0.037 | 0.061 | 0.063 | 0.046 | 0.068 | 0.067 | 0.058 | 0.092 | 0.099 | 0.076 | 0.120 | 0.110 | 0.093 | 0.152 |
| LR | 0.037 | 0.029 | 0.044 | 0.044 | 0.036 | 0.058 | 0.074 | 0.057 | 0.094 | 0.094 | 0.078 | 0.128 | 0.239 | 0.183 | 0.298 |
| XGBoost | 0.052 | 0.033 | 0.050 | 0.068 | 0.048 | 0.073 | 0.063 | 0.045 | 0.069 | 0.094 | 0.073 | 0.119 | **0.090** | **0.071** | **0.112** |
| RT | 0.063 | 0.042 | 0.065 | 0.073 | 0.054 | 0.081 | 0.082 | 0.063 | 0.095 | 0.106 | 0.084 | 0.136 | 0.099 | 0.077 | 0.122 |
| PM | 0.039 | 0.025 | 0.037 | 0.082 | 0.059 | 0.087 | 0.110 | 0.087 | 0.130 | 0.078 | 0.057 | 0.089 | 0.126 | 0.100 | 0.149 |
| PatchTST | 0.091 | 0.074 | 0.101 | 0.066 | 0.053 | 0.076 | 0.063 | 0.052 | 0.080 | 0.080 | 0.068 | 0.109 | 0.140 | 0.114 | 0.186 |
| TimeLLM | 0.137 | 0.123 | 0.176 | 0.286 | 0.236 | 0.371 | 0.232 | 0.191 | 0.305 | 0.242 | 0.189 | 0.312 | 0.167 | 0.134 | 0.224 |

TABLE IV

THE RESULTS OF MACHINE LEARNING MODELS IN LOAD FORECASTING WITH A SHORT LOOK-AHEAD TIME (CASE 2)

| Model | Look-ahead time=1 hour | | | Look-ahead time=4 hours | | | Look-ahead time=6 hours | | | Look-ahead time=12 hours | | | Look-ahead time=24 hours | | |
|---|---|---|---|---|---|---|---|---|---|---|---|---|---|---|---|
| | RMSE | MAE | MAPE | RMSE | MAE | MAPE | RMSE | MAE | MAPE | RMSE | MAE | MAPE | RMSE | MAE | MAPE |
| TimeGPT | **0.021** | **0.015** | **0.022** | **0.049** | **0.035** | **0.053** | **0.065** | **0.043** | **0.067** | **0.079** | **0.059** | **0.099** | 0.115 | 0.087 | 0.133 |
| MLP | 0.030 | 0.026 | 0.038 | 0.073 | 0.052 | 0.083 | 0.080 | 0.061 | 0.100 | 0.092 | 0.075 | 0.129 | 0.102 | 0.081 | 0.141 |
| LSTM | 0.022 | 0.019 | 0.025 | 0.076 | 0.050 | 0.080 | 0.097 | 0.076 | 0.126 | 0.096 | 0.075 | 0.129 | 0.102 | 0.082 | 0.141 |
| LR | 0.022 | 0.016 | 0.024 | 0.053 | 0.037 | 0.059 | 0.070 | 0.054 | 0.088 | 0.093 | 0.077 | 0.133 | 0.101 | 0.079 | 0.139 |
| XGBoost | 0.032 | 0.028 | 0.038 | 0.072 | 0.054 | 0.085 | 0.080 | 0.061 | 0.099 | 0.098 | 0.077 | 0.132 | 0.099 | 0.075 | 0.130 |
| RT | 0.033 | 0.029 | 0.040 | 0.075 | 0.054 | 0.088 | 0.086 | 0.060 | 0.096 | 0.108 | 0.083 | 0.144 | 0.108 | 0.081 | 0.140 |
| PM | 0.038 | 0.028 | 0.039 | 0.070 | 0.050 | 0.076 | 0.081 | 0.056 | 0.088 | 0.082 | 0.063 | 0.106 | 0.116 | 0.092 | 0.145 |
| PatchTST | 0.021 | 0.016 | 0.023 | 0.066 | 0.049 | 0.081 | 0.073 | 0.053 | 0.088 | 0.081 | 0.061 | 0.102 | **0.096** | **0.075** | **0.125** |
| TimeLLM | 0.126 | 0.112 | 0.177 | 0.330 | 0.271 | 0.432 | 0.264 | 0.209 | 0.346 | 0.213 | 0.166 | 0.281 | 0.257 | 0.207 | 0.354 |

TABLE V

THE RESULTS OF MACHINE LEARNING MODELS IN LOAD FORECASTING WITH A SHORT LOOK-AHEAD TIME (CASE 3)

| Model | Look-ahead time=1 hour | | | Look-ahead time=4 hours | | | Look-ahead time=6 hours | | | Look-ahead time=12 hours | | | Look-ahead time=24 hours | | |
|---|---|---|---|---|---|---|---|---|---|---|---|---|---|---|---|
| | RMSE | MAE | MAPE | RMSE | MAE | MAPE | RMSE | MAE | MAPE | RMSE | MAE | MAPE | RMSE | MAE | MAPE |
| TimeGPT | **0.016** | **0.012** | **0.021** | **0.055** | **0.036** | **0.061** | 0.088 | 0.059 | 0.102 | 0.083 | 0.063 | 0.106 | 0.132 | 0.102 | 0.157 |
| MLP | 0.028 | 0.022 | 0.038 | 0.057 | 0.049 | 0.088 | 0.063 | 0.054 | 0.096 | 0.081 | 0.068 | 0.117 | 0.100 | 0.081 | 0.146 |
| LSTM | 0.039 | 0.035 | 0.061 | 0.089 | 0.076 | 0.138 | 0.093 | 0.078 | 0.142 | 0.108 | 0.091 | 0.156 | 0.120 | 0.096 | 0.172 |
| LR | 0.019 | 0.014 | 0.024 | 0.055 | 0.046 | 0.081 | 0.068 | 0.057 | 0.102 | 0.082 | 0.069 | 0.118 | 0.099 | 0.078 | 0.141 |
| XGBoost | 0.022 | 0.017 | 0.030 | 0.080 | 0.062 | 0.113 | 0.087 | 0.070 | 0.128 | 0.100 | 0.082 | 0.143 | 0.108 | 0.085 | 0.151 |
| RT | 0.030 | 0.023 | 0.041 | 0.101 | 0.079 | 0.144 | 0.091 | 0.074 | 0.133 | 0.112 | 0.085 | 0.146 | 0.123 | 0.096 | 0.170 |
| PM | 0.017 | 0.013 | 0.022 | 0.058 | 0.038 | 0.064 | 0.098 | 0.070 | 0.121 | 0.091 | 0.071 | 0.120 | 0.132 | 0.105 | 0.166 |
| PatchTST | 0.017 | 0.012 | 0.022 | 0.057 | 0.038 | 0.063 | **0.062** | **0.045** | **0.072** | **0.073** | **0.052** | **0.088** | **0.093** | **0.071** | **0.115** |
| TimeLLM | 0.046 | 0.038 | 0.064 | 0.151 | 0.110 | 0.197 | 0.112 | 0.092 | 0.163 | 0.230 | 0.188 | 0.313 | 0.196 | 0.162 | 0.282 |

***1) Performance Comparison***

In scenarios with scarce historical data, TimeGPT demonstrates significant advantages over benchmarks for load forecasting, particularly with a look-ahead time of a few hours (e.g., 1 hour to 6 hours). For instance, in case 1 with a 1-hour look-ahead time, the RMSE of TimeGPT is reduced by 35.29%, 23.26%, 10.81%, 36.54%, 47.62%, 15.38%, 63.73%, and 75.91%, compared to MLP, LSTM, LR, XGBoost, RT, PM, PatchTST, and TimeLLM, respectively. The likely reasons for the strong performance of TimeGPT in load forecasting are as follows:

TimeGPT benefits from pre-training on massive and diverse time series datasets, which gives it a degree of generalization. Even in scenarios with scarce data, it can use this rich prior knowledge to perform well on load forecasting. In contrast, traditional machine learning models struggle to capture complex patterns in load data due to insufficient training data, resulting in lower forecast accuracy.

Similarly, in Case 2 (5 days of training data), TimeGPT continues to outperform baselines for load forecasting with a look-ahead time of 1 to 12 hours. In Case 3 (7 days of training data), TimeGPT still outperforms the baselines but with a narrower range of the look-ahead time. However, in Cases 4 (15 days of training data) and 5 (30 days of training data), where more historical data is available, the performance advantages of TimeGPT diminish and it may underperform the baselines for load forecasting. These observations suggest that while TimeGPT performs well with extremely limited historical data, its advantages become less apparent as the amount of training data increases. This highlights the strength of TimeGPT in dealing with data scarcity, especially when the historical data is less than 7 days old.

In Tables III to V, TimeLLM shows poor performance. This is mainly due to the fact that it is trained on large amounts of text data, making it less suited for time series tasks. In contrast, TimeGPT is trained directly on time series data. Its parameters are optimized for temporal patterns, allowing it to handle load forecasting tasks with greater accuracy and efficiency.

However, the day-ahead load forecasting (e.g., the look-ahead time is 24 hours) involves longer temporal dependencies, which may exceed the scope of what TimeGPT learned during pre-training. As a result, its performance may be relatively worse compared to the benchmarks. Similarly, TimeGPT does not perform well in load forecasting where the look-ahead time is longer (e.g. the look-ahead time is greater than 24 hours).

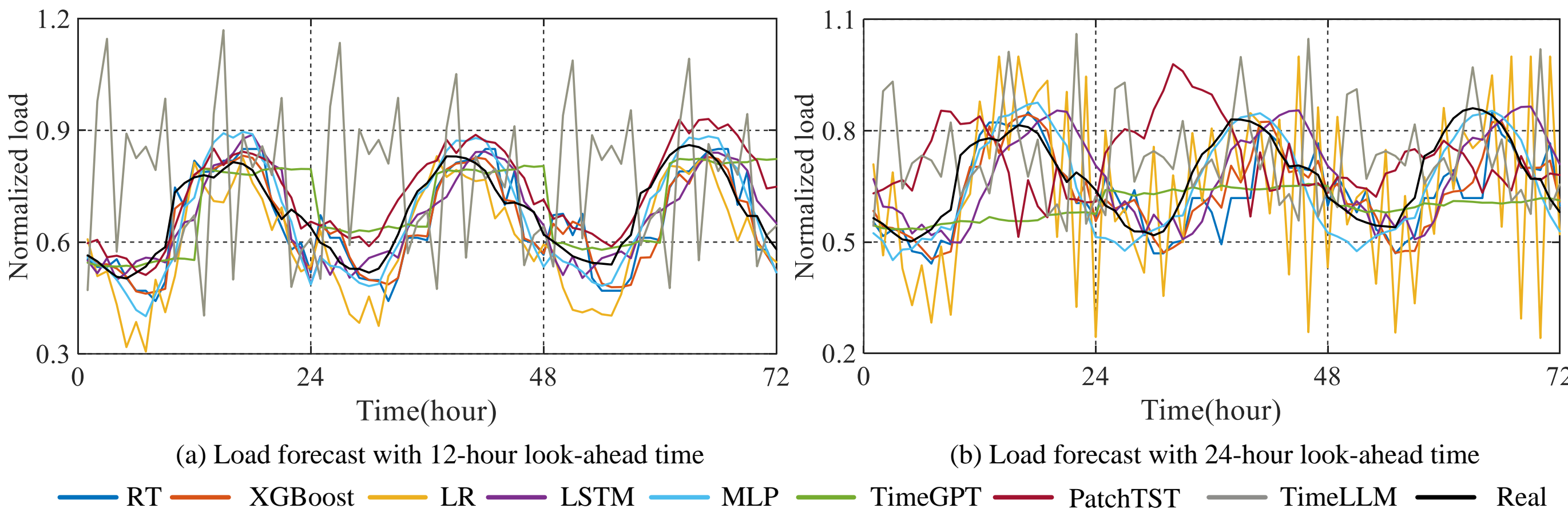

Fig. 6. The load forecasting of TimeGPT with a long look-ahead time (e.g., 12 hours and 24 hours).

TABLE VI

THE RESULTS OF MACHINE LEARNING MODELS IN LOAD FORECASTING WITH A LONG LOOK-AHEAD TIME (CASE 4)

| Model | Look-ahead time=1 hour | | | Look-ahead time=4 hours | | | Look-ahead time=6 hours | | | Look-ahead time=12 hours | | | Look-ahead time=24 hours | | |
|---|---|---|---|---|---|---|---|---|---|---|---|---|---|---|---|
| | RMSE | MAE | MAPE | RMSE | MAE | MAPE | RMSE | MAE | MAPE | RMSE | MAE | MAPE | RMSE | MAE | MAPE |
| TimeGPT | 0.048 | 0.034 | 0.055 | 0.082 | 0.054 | 0.084 | 0.117 | 0.087 | 0.138 | 0.107 | 0.082 | 0.130 | 0.134 | 0.102 | 0.148 |
| MLP | 0.029 | 0.025 | 0.042 | 0.048 | 0.038 | 0.065 | **0.053** | 0.044 | 0.075 | 0.051 | 0.041 | 0.071 | **0.065** | **0.049** | **0.075** |
| LSTM | **0.027** | **0.024** | 0.039 | 0.046 | 0.038 | 0.065 | 0.059 | 0.051 | 0.090 | 0.052 | 0.042 | 0.073 | 0.067 | 0.052 | 0.082 |
| LR | 0.029 | **0.024** | **0.038** | **0.041** | **0.030** | **0.049** | **0.053** | **0.038** | **0.061** | **0.042** | **0.032** | **0.051** | 0.067 | 0.052 | 0.081 |
| XGBoost | 0.035 | 0.031 | 0.055 | 0.064 | 0.051 | 0.098 | 0.070 | 0.056 | 0.103 | 0.078 | 0.052 | 0.093 | 0.080 | 0.056 | 0.091 |
| RT | 0.041 | 0.034 | 0.056 | 0.099 | 0.067 | 0.128 | 0.105 | 0.077 | 0.142 | 0.101 | 0.070 | 0.121 | 0.092 | 0.065 | 0.105 |
| PM | 0.036 | 0.027 | 0.043 | 0.089 | 0.060 | 0.095 | 0.134 | 0.103 | 0.168 | 0.099 | 0.079 | 0.131 | 0.126 | 0.097 | 0.145 |
| PatchTST | 0.045 | 0.035 | 0.053 | 0.052 | 0.040 | 0.062 | 0.065 | 0.047 | 0.073 | 0.049 | 0.035 | 0.054 | 0.079 | 0.056 | 0.084 |
| TimeLLM | 0.243 | 0.196 | 0.291 | 0.209 | 0.167 | 0.291 | 0.210 | 0.174 | 0.302 | 0.260 | 0.209 | 0.327 | 0.186 | 0.156 | 0.257 |

TABLE VII

THE RESULTS OF MACHINE LEARNING MODELS IN LOAD FORECASTING WITH A LONG LOOK-AHEAD TIME (CASE 5)

| Model | Look-ahead time=1 hour | | | Look-ahead time=4 hours | | | Look-ahead time=6 hours | | | Look-ahead time=12 hours | | | Look-ahead time=24 hours | | |
|---|---|---|---|---|---|---|---|---|---|---|---|---|---|---|---|
| | RMSE | MAE | MAPE | RMSE | MAE | MAPE | RMSE | MAE | MAPE | RMSE | MAE | MAPE | RMSE | MAE | MAPE |
| TimeGPT | 0.028 | 0.022 | 0.036 | 0.082 | 0.059 | 0.098 | 0.113 | 0.081 | 0.139 | 0.091 | 0.070 | 0.119 | 0.149 | 0.122 | 0.176 |
| MLP | 0.039 | 0.035 | 0.056 | 0.099 | 0.081 | 0.132 | 0.138 | 0.116 | 0.193 | 0.132 | 0.106 | 0.174 | 0.153 | 0.127 | 0.186 |
| LSTM | 0.016 | 0.013 | 0.020 | **0.029** | **0.023** | **0.038** | **0.036** | **0.029** | **0.052** | **0.043** | 0.036 | **0.061** | **0.060** | **0.046** | **0.070** |
| LR | **0.012** | **0.009** | **0.016** | 0.033 | 0.026 | 0.043 | 0.037 | 0.029 | 0.051 | 0.056 | 0.047 | 0.079 | 0.078 | 0.055 | 0.083 |
| XGBoost | 0.020 | 0.015 | 0.024 | 0.037 | 0.030 | 0.050 | 0.045 | 0.035 | 0.061 | 0.044 | **0.035** | **0.061** | 0.065 | 0.049 | 0.076 |
| RT | 0.021 | 0.016 | 0.026 | 0.029 | 0.022 | 0.037 | 0.041 | 0.029 | 0.053 | 0.045 | 0.035 | 0.059 | 0.063 | 0.047 | 0.071 |
| PM | 0.025 | 0.021 | 0.033 | 0.047 | 0.037 | 0.061 | 0.056 | 0.041 | 0.069 | 0.058 | 0.046 | 0.077 | 0.083 | 0.062 | 0.094 |
| PatchTST | 0.017 | 0.013 | 0.020 | 0.031 | 0.025 | 0.042 | 0.041 | 0.032 | 0.057 | 0.051 | 0.039 | 0.067 | 0.070 | 0.053 | 0.083 |
| TimeLLM | 0.158 | 0.134 | 0.194 | 0.228 | 0.183 | 0.303 | 0.187 | 0.153 | 0.275 | 0.184 | 0.152 | 0.269 | 0.240 | 0.194 | 0.316 |

#### *2) Visual Analysis on a Long Look-Ahead Time*

To explore why TimeGPT is not efficient in load forecasting with a long look-ahead time (e.g., 12 hours and 24 hours), a specific case (i.e., Case 1) is selected as an example to visualize load forecasting using the recursive forecasting method. Specifically, three days of load data are randomly selected to perform load forecasting with a long look-ahead time (e.g., 12 hours and 24 hours), as shown in Fig. 6. Note that PM is removed because it is not suitable for load forecasting with a long look-ahead time.

Visual analysis shows that the forecasts generated by TimeGPT have a conservative and smoothed pattern. This pattern makes it difficult for TimeGPT to capture the peaks and valleys of load, which explains its low accuracy in load forecasting with a long look-ahead time (e.g., 12 hours and 24 hours).

### *D. Performance Comparison in Data Rich Cases*

To compare the performance of TimeGPT with the benchmarks in cases where data is relatively rich, these models are used to conduct load forecasting in Cases 4-5. Note that TimeGPT is fine-tuned by the historical load data here. The average evaluation metrics of TimeGPT in various scenarios are shown in Tables VI-VII.

Tables VI and VII show that, despite the fine-tuning, TimeGPT performs significantly worse than machine learning models in load forecasting with relatively rich historical data. This could be due to potential mismatches in distribution and characteristics between the training dataset and the load data. Machine learning models trained directly on load data have an optimization process entirely focused on load forecasting, allowing them to better adapt to the specific characteristics of load forecasting. In contrast, when the pre-training data of TimeGPT comes from different time series, such as traffic, weather, energy, network and financial data, the learned representations may capture a wide range of temporal patterns that do not exactly match the unique characteristics of the load forecast. The fine-tuning process, while beneficial, may not be sufficient to overcome these underlying differences as the pre-training weights are initially optimized for generic tasks rather than the specific nuances of load forecasting. In other words, if sufficient data is available to train machine learning models, the choice of classical machine learning models may be more desirable than LTSMs like TimeGPT.

The impact of such distributional differences is twofold: firstly, it may lead to suboptimal feature extraction during the fine-tuning phase, as the internal representations in TimeGPT may be biased towards patterns that are prevalent in the pre-training data but less relevant to load forecasting. Secondly, these differences can cause TimeGPT to struggle to capture the intricate seasonal and temporal dependencies unique to load data, leading to poor performance.

To mitigate these challenges, several strategies can be considered in future work. One possible research line is to incorporate domain adaptation techniques during the fine-tuning phase (e.g., adversarial training) that emphasize alignment between the pre-training and target data distributions (i.e., load data). Another research line could be the use of a hybrid model that combines TimeGPT with a traditional load forecasting model, exploiting the strengths of both. Additionally, the expansion of the fine-tuning dataset to include more diverse load-related examples or the use of techniques such as data augmentation could also help narrow the distributional gap.

# IV. Case Study II

This section will further investigate the generalization of TimeGPT for other load datasets.

## A. Simulation Settings

The simulations are performed on four publicly available datasets from China Nongfu Spring Company (a packaged water supplier) [40], Midea Group (an electrical appliance manufacturer) [40], the Joho City Electric Power Company in Malaysia [41], and Arizona State University Tempe Campus [42], respectively. The time resolution in these four datasets is one hour. Although the first two datasets are publicly available, they should not have been used to train TimeGPT, because a password is required to obtain them.

Due to limited API tokens for TimeGPT, the three months of load data are used for simulation and testing. Specifically, the dataset from Nongfu Spring Company spans from May 1, 2017, to July 31, 2017. The dataset from Midea Group covers the period from April 28, 2017, to July 27, 2017. Meanwhile, the dataset from Joho City ranges from January 1, 2009, to March 31, 2009. The dataset from Arizona State University ranges from January 1, 2012, to March 31, 2012.

## B. Results and Analysis

Similar to Section III, the TimeGPT and benchmarks are used to conduct load forecasting in the data scarce case (e.g., Case 1) and data rich case (e.g., Case 5). The parameter settings for the different cases are the same as before:

Case 1: The first 3 days of load data are used as the training set, and the remainder is considered the test set.

Case 5: The first 30 days of load data are used as the training set, and the remainder is considered the test set.

The average evaluation metrics in data rich cases are presented in Tables VIII-XI, and the average evaluation metrics in data scarce cases are presented in Tables XII-XV.

TABLE VIII

RESULTS IN CHINA NONGFU SPRING COMPANY'S DATASET IN DATA RICH CASE (I.E., CASE 5)

| Model | Look-ahead time=1 hour | | | Look-ahead time=4 hours | | | Look-ahead time=6 hours | | | Look-ahead time=12 hours | | | Look-ahead time=24 hours | | |
|---|---|---|---|---|---|---|---|---|---|---|---|---|---|---|---|
| | RMSE | MAE | MAPE | RMSE | MAE | MAPE | RMSE | MAE | MAPE | RMSE | MAE | MAPE | RMSE | MAE | MAPE |
| TimeGPT | 0.059 | 0.046 | 0.107 | 0.105 | 0.080 | 0.147 | 0.140 | 0.102 | 0.183 | 0.162 | 0.112 | 0.173 | 0.148 | 0.112 | 0.175 |
| MLP | 0.056 | 0.042 | 0.098 | 0.114 | 0.092 | 0.151 | 0.142 | 0.112 | 0.201 | 0.151 | 0.115 | 0.169 | 0.150 | 0.118 | 0.168 |
| LSTM | 0.055 | 0.044 | 0.101 | 0.112 | 0.095 | 0.168 | 0.150 | 0.119 | 0.209 | 0.183 | 0.147 | 0.210 | 0.152 | 0.116 | 0.171 |
| LR | **0.053** | **0.041** | **0.100** | **0.094** | **0.073** | **0.127** | **0.113** | **0.083** | **0.149** | **0.126** | **0.090** | **0.138** | **0.130** | **0.105** | **0.150** |
| XGBoost | 0.062 | 0.048 | 0.113 | 0.114 | 0.098 | 0.163 | 0.153 | 0.131 | 0.216 | 0.141 | 0.107 | 0.159 | 0.140 | 0.108 | 0.153 |
| RT | 0.068 | 0.057 | 0.128 | 0.115 | 0.089 | 0.141 | 0.172 | 0.141 | 0.219 | 0.173 | 0.136 | 0.203 | 0.183 | 0.138 | 0.206 |
| PM | 0.060 | 0.047 | 0.112 | 0.108 | 0.077 | 0.146 | 0.137 | 0.100 | 0.183 | 0.161 | 0.104 | 0.163 | 0.145 | 0.109 | 0.173 |
| PatchTST | 0.058 | 0.045 | 0.105 | 0.114 | 0.092 | 0.151 | 0.126 | 0.095 | 0.160 | 0.152 | 0.112 | 0.176 | 0.138 | 0.104 | 0.157 |
| TimeLLM | 0.197 | 0.159 | 0.306 | 0.286 | 0.226 | 0.357 | 0.226 | 0.178 | 0.259 | 0.246 | 0.189 | 0.279 | 0.273 | 0.215 | 0.300 |

TABLE IX

RESULTS IN MIDEA GROUP'S DATASET IN DATA RICH CASE (I.E., CASE 5)

| Model | Look-ahead time=1 hour | | | Look-ahead time=4 hours | | | Look-ahead time=6 hours | | | Look-ahead time=12 hours | | | Look-ahead time=24 hours | | |
|---|---|---|---|---|---|---|---|---|---|---|---|---|---|---|---|
| | RMSE | MAE | MAPE | RMSE | MAE | MAPE | RMSE | MAE | MAPE | RMSE | MAE | MAPE | RMSE | MAE | MAPE |
| TimeGPT | 0.074 | 0.057 | 0.130 | 0.185 | 0.131 | 0.398 | 0.178 | 0.136 | 0.747 | 0.199 | 0.157 | 0.775 | 0.272 | 0.228 | **1.111** |
| MLP | 0.062 | 0.042 | 0.094 | **0.084** | **0.059** | **0.185** | **0.113** | **0.077** | 0.800 | 0.215 | 0.150 | 2.933 | 0.210 | **0.133** | 2.022 |
| LSTM | 0.049 | 0.038 | 0.098 | 0.121 | 0.081 | 0.314 | 0.149 | 0.112 | 0.849 | 0.283 | 0.213 | 4.051 | 0.272 | 0.191 | 2.387 |
| LR | **0.043** | 0.037 | 0.092 | 0.089 | 0.069 | 0.222 | 0.145 | 0.091 | 1.177 | **0.161** | 0.116 | 1.971 | **0.202** | 0.142 | 1.783 |
| XGBoost | 0.047 | **0.036** | **0.088** | 0.085 | 0.064 | 0.220 | 0.143 | 0.088 | 1.125 | 0.241 | 0.174 | 3.620 | 0.230 | 0.163 | 2.202 |
| RT | 0.071 | 0.053 | 0.127 | 0.133 | 0.088 | 0.282 | 0.178 | 0.114 | 1.284 | 0.250 | 0.181 | 3.224 | 0.245 | 0.171 | 2.101 |
| PM | 0.090 | 0.071 | 0.159 | 0.214 | 0.150 | 0.431 | 0.203 | 0.159 | **0.723** | 0.188 | **0.149** | **0.769** | 0.283 | 0.234 | 1.194 |
| PatchTST | 0.060 | 0.048 | 0.137 | 0.090 | 0.068 | 0.269 | 0.149 | 0.092 | 1.206 | 0.201 | 0.152 | 1.886 | 0.228 | 0.152 | 1.565 |
| TimeLLM | 0.206 | 0.189 | 0.554 | 0.294 | 0.232 | 1.037 | 0.397 | 0.303 | 2.354 | 0.365 | 0.309 | 3.867 | 0.552 | 0.443 | 3.043 |

TABLE X

RESULTS IN JOHO CITY'S DATASET IN DATA RICH CASE (I.E., CASE 5)

| Model | Look-ahead time=1 hour | | | Look-ahead time=4 hours | | | Look-ahead time=6 hours | | | Look-ahead time=12 hours | | | Look-ahead time=24 hours | | |
|---|---|---|---|---|---|---|---|---|---|---|---|---|---|---|---|
| | RMSE | MAE | MAPE | RMSE | MAE | MAPE | RMSE | MAE | MAPE | RMSE | MAE | MAPE | RMSE | MAE | MAPE |
| TimeGPT | 0.054 | 0.044 | 0.203 | 0.118 | 0.082 | 0.330 | 0.187 | 0.119 | 0.440 | 0.241 | 0.162 | 0.531 | 0.312 | 0.237 | 0.569 |
| MLP | **0.020** | **0.015** | 0.068 | **0.036** | **0.026** | 0.142 | 0.064 | **0.040** | **0.152** | **0.055** | **0.030** | **0.106** | **0.087** | 0.059 | 0.196 |
| LSTM | 0.022 | 0.017 | 0.068 | 0.038 | 0.031 | 0.170 | **0.061** | **0.040** | 0.157 | 0.065 | 0.039 | 0.134 | 0.096 | 0.068 | 0.231 |
| LR | 0.039 | 0.031 | 0.128 | 0.077 | 0.050 | 0.195 | 0.112 | 0.068 | 0.228 | 0.084 | 0.049 | 0.162 | 0.095 | 0.060 | 0.185 |
| XGBoost | 0.031 | 0.023 | **0.102** | 0.048 | 0.031 | **0.126** | 0.088 | 0.055 | 0.174 | 0.062 | 0.035 | 0.113 | 0.090 | 0.058 | **0.160** |
| RT | 0.037 | 0.024 | 0.127 | 0.042 | **0.026** | 0.133 | 0.104 | 0.053 | 0.185 | 0.076 | 0.036 | 0.128 | 0.104 | **0.057** | 0.170 |
| PM | 0.056 | 0.038 | 0.158 | 0.144 | 0.096 | 0.424 | 0.203 | 0.136 | 0.594 | 0.243 | 0.171 | 0.673 | 0.286 | 0.224 | 0.643 |
| PatchTST | 0.030 | 0.023 | 0.090 | 0.034 | 0.023 | 0.121 | 0.106 | 0.059 | 0.232 | 0.090 | 0.056 | 0.186 | 0.105 | 0.072 | 0.244 |
| TimeLLM | 0.274 | 0.217 | 0.600 | 0.273 | 0.226 | 1.603 | 0.458 | 0.366 | 2.464 | 0.450 | 0.362 | 2.203 | 0.376 | 0.274 | 1.545 |

TABLE XI

RESULTS IN ARIZONA STATE UNIVERSITY'S DATASET IN DATA RICH CASE (I.E., CASE 5)

| Model | Look-ahead time=1 hour | | | Look-ahead time=4 hours | | | Look-ahead time=6 hours | | | Look-ahead time=12 hours | | | Look-ahead time=24 hours | | |
|---|---|---|---|---|---|---|---|---|---|---|---|---|---|---|---|
| | RMSE | MAE | MAPE | RMSE | MAE | MAPE | RMSE | MAE | MAPE | RMSE | MAE | MAPE | RMSE | MAE | MAPE |
| TimeGPT | 0.032 | 0.024 | 0.087 | 0.126 | 0.086 | 0.290 | 0.200 | 0.152 | 0.377 | 0.176 | 0.130 | 0.358 | 0.238 | 0.185 | 0.471 |
| MLP | 0.019 | 0.014 | 0.038 | 0.041 | 0.031 | 0.075 | 0.053 | 0.041 | 0.093 | **0.079** | **0.065** | **0.150** | 0.118 | 0.089 | 0.201 |
| LSTM | 0.015 | 0.011 | **0.032** | 0.034 | **0.026** | 0.068 | **0.047** | **0.034** | **0.075** | 0.107 | 0.078 | 0.162 | **0.110** | **0.079** | **0.200** |
| LR | 0.021 | 0.017 | 0.048 | 0.059 | 0.048 | 0.114 | 0.086 | 0.063 | 0.138 | 0.093 | 0.068 | 0.169 | 0.128 | 0.095 | 0.231 |
| XGBoost | **0.014** | **0.010** | 0.034 | 0.034 | 0.026 | **0.063** | 0.072 | 0.049 | 0.097 | 0.087 | 0.068 | 0.163 | 0.117 | 0.087 | 0.205 |
| RT | 0.023 | 0.017 | 0.060 | **0.033** | 0.026 | 0.075 | 0.093 | 0.060 | 0.140 | 0.105 | 0.081 | 0.202 | 0.142 | 0.101 | 0.232 |
| PM | 0.052 | 0.039 | 0.135 | 0.135 | 0.106 | 0.356 | 0.242 | 0.200 | 0.570 | 0.155 | 0.126 | 0.442 | 0.218 | 0.175 | 0.539 |
| PatchTST | 0.018 | 0.012 | 0.041 | 0.048 | 0.033 | 0.091 | 0.051 | 0.037 | 0.096 | 0.078 | 0.059 | 0.184 | 0.135 | 0.101 | 0.251 |
| TimeLLM | 0.254 | 0.219 | 0.493 | 0.430 | 0.356 | 1.242 | 0.285 | 0.235 | 0.819 | 0.337 | 0.268 | 0.987 | 0.250 | 0.203 | 0.853 |

TABLE XII

RESULTS IN CHINA NONGFU SPRING COMPANY'S DATASET IN DATA SCARCE CASE (I.E., CASE 1)

| Model | Look-ahead time=1 hour | | | Look-ahead time=4 hours | | | Look-ahead time=6 hours | | | Look-ahead time=12 hours | | | Look-ahead time=24 hours | | |
|---|---|---|---|---|---|---|---|---|---|---|---|---|---|---|---|
| | RMSE | MAE | MAPE | RMSE | MAE | MAPE | RMSE | MAE | MAPE | RMSE | MAE | MAPE | RMSE | MAE | MAPE |
| TimeGPT | **0.047** | **0.037** | **0.073** | **0.088** | **0.072** | **0.127** | **0.141** | **0.091** | **0.186** | **0.125** | **0.096** | **0.215** | **0.146** | **0.108** | **0.294** |
| MLP | 0.116 | 0.094 | 0.199 | 0.161 | 0.129 | 0.221 | 0.218 | 0.189 | 0.307 | 0.255 | 0.208 | 0.403 | 0.239 | 0.201 | 0.496 |
| LSTM | 0.096 | 0.079 | 0.166 | 0.228 | 0.177 | 0.293 | 0.319 | 0.261 | 0.403 | 0.265 | 0.214 | 0.446 | 0.362 | 0.312 | 0.716 |
| LR | 0.084 | 0.071 | 0.141 | 0.168 | 0.135 | 0.235 | 0.212 | 0.162 | 0.268 | 0.324 | 0.245 | 0.476 | 0.582 | 0.432 | 0.960 |
| XGBoost | 0.062 | 0.047 | 0.088 | 0.198 | 0.157 | 0.259 | 0.275 | 0.223 | 0.339 | 0.237 | 0.179 | 0.338 | 0.224 | 0.180 | 0.427 |
| RT | 0.060 | 0.050 | 0.101 | 0.204 | 0.163 | 0.278 | 0.273 | 0.227 | 0.352 | 0.261 | 0.200 | 0.389 | 0.221 | 0.177 | 0.428 |
| PM | 0.053 | 0.042 | 0.084 | 0.091 | 0.070 | 0.123 | 0.147 | 0.096 | 0.198 | 0.136 | 0.101 | 0.234 | 0.165 | 0.120 | 0.333 |
| PatchTST | 0.102 | 0.089 | 0.191 | 0.169 | 0.139 | 0.271 | 0.160 | 0.116 | 0.245 | 0.160 | 0.120 | 0.287 | 0.264 | 0.200 | 0.544 |
| TimeLLM | 0.147 | 0.122 | 0.225 | 0.307 | 0.282 | 0.523 | 0.295 | 0.255 | 0.473 | 0.279 | 0.218 | 0.520 | 0.286 | 0.215 | 0.526 |

TABLE XIII

RESULTS IN MIDEA GROUP'S DATASET IN DATA SCARCE CASE (I.E., CASE 1)

| Model | Look-ahead time=1 hour | | | Look-ahead time=4 hours | | | Look-ahead time=6 hours | | | Look-ahead time=12 hours | | | Look-ahead time=24 hours | | |
|---|---|---|---|---|---|---|---|---|---|---|---|---|---|---|---|
| | RMSE | MAE | MAPE | RMSE | MAE | MAPE | RMSE | MAE | MAPE | RMSE | MAE | MAPE | RMSE | MAE | MAPE |
| TimeGPT | **0.051** | **0.037** | **0.170** | 0.185 | 0.117 | 0.650 | 0.204 | 0.151 | 0.750 | **0.167** | **0.134** | 0.647 | **0.234** | **0.195** | 0.627 |
| MLP | 0.070 | 0.056 | 0.234 | 0.116 | **0.084** | **0.366** | 0.174 | 0.136 | **0.460** | 0.345 | 0.294 | 0.638 | 0.343 | 0.297 | 0.597 |
| LSTM | 0.117 | 0.100 | 0.389 | 0.206 | 0.164 | 1.099 | 0.198 | 0.147 | 0.562 | 0.263 | 0.218 | **0.479** | 0.381 | 0.313 | 0.641 |
| LR | 0.051 | 0.038 | 0.201 | 0.137 | 0.109 | 0.815 | **0.151** | **0.117** | 0.616 | 0.376 | 0.290 | 0.808 | 0.434 | 0.345 | 0.703 |
| XGBoost | 0.090 | 0.076 | 0.385 | 0.144 | 0.108 | 0.522 | 0.207 | 0.153 | 0.462 | 0.331 | 0.278 | 0.555 | 0.364 | 0.307 | **0.574** |
| RT | 0.138 | 0.104 | 0.589 | 0.158 | 0.113 | 0.487 | 0.226 | 0.173 | 0.456 | 0.328 | 0.276 | 0.561 | 0.372 | 0.314 | 0.618 |
| PM | 0.063 | 0.036 | 0.162 | 0.203 | 0.126 | 0.736 | 0.223 | 0.164 | 0.829 | 0.169 | 0.146 | 0.674 | 0.247 | 0.198 | 0.654 |
| PatchTST | 0.086 | 0.073 | 0.279 | 0.205 | 0.174 | 1.223 | 0.247 | 0.216 | 1.040 | 0.258 | 0.231 | 0.896 | 0.267 | 0.218 | 0.837 |
| TimeLLM | 0.176 | 0.137 | 0.394 | 0.592 | 0.489 | 3.438 | 0.268 | 0.218 | 0.991 | 0.280 | 0.224 | 0.858 | 0.250 | 0.197 | 0.726 |

TABLE XIV

RESULTS IN JOHO CITY'S DATASET IN DATA SCARCE CASE (I.E., CASE 1)

| Model | Look-ahead time=1 hour | | | Look-ahead time=4 hours | | | Look-ahead time=6 hours | | | Look-ahead time=12 hours | | | Look-ahead time=24 hours | | |
|---|---|---|---|---|---|---|---|---|---|---|---|---|---|---|---|
| | RMSE | MAE | MAPE | RMSE | MAE | MAPE | RMSE | MAE | MAPE | RMSE | MAE | MAPE | RMSE | MAE | MAPE |
| TimeGPT | **0.019** | **0.015** | **0.078** | **0.104** | **0.081** | **0.299** | **0.153** | 0.134 | **0.417** | 0.269 | 0.181 | 0.552 | 0.341 | 0.259 | 0.557 |
| MLP | 0.080 | 0.070 | 0.438 | 0.141 | 0.124 | 0.644 | 0.154 | 0.130 | 0.576 | **0.129** | 0.096 | 0.328 | **0.146** | 0.117 | **0.453** |
| LSTM | 0.104 | 0.084 | 0.735 | 0.294 | 0.272 | 1.715 | 0.224 | 0.184 | 1.054 | 0.185 | 0.136 | 0.561 | 0.193 | 0.147 | 0.698 |
| LR | 0.053 | 0.038 | 0.141 | 0.165 | 0.105 | 0.339 | 0.219 | **0.112** | 0.431 | 0.583 | 0.417 | 1.823 | 0.864 | 0.554 | 2.065 |
| XGBoost | 0.084 | 0.072 | 0.580 | 0.181 | 0.141 | 0.664 | 0.155 | 0.117 | 0.543 | 0.163 | **0.114** | 0.487 | 0.163 | **0.115** | **0.407** |
| RT | 0.126 | 0.115 | 0.806 | 0.241 | 0.191 | 0.857 | 0.201 | 0.146 | 0.705 | 0.215 | 0.138 | 0.718 | 0.205 | 0.147 | 0.530 |
| PM | 0.051 | 0.030 | 0.192 | 0.178 | 0.107 | 0.382 | 0.246 | 0.155 | 0.575 | 0.272 | 0.191 | 0.684 | 0.324 | 0.254 | 0.655 |
| PatchTST | 0.098 | 0.082 | 0.519 | 0.155 | 0.117 | 0.412 | 0.164 | 0.123 | 0.488 | 0.153 | 0.098 | 0.374 | 0.223 | 0.167 | 0.591 |
| TimeLLM | 0.290 | 0.243 | 0.838 | 0.558 | 0.448 | 2.304 | 0.328 | 0.270 | 1.661 | 0.337 | 0.278 | 1.526 | 0.380 | 0.333 | 1.735 |

TABLE XV

RESULTS IN ARIZONA STATE UNIVERSITY'S DATASET IN DATA SCARCE CASE (I.E., CASE 1)

| Model | Look-ahead time=1 hour | | | Look-ahead time=4 hours | | | Look-ahead time=6 hours | | | Look-ahead time=12 hours | | | Look-ahead time=24 hours | | |
|---|---|---|---|---|---|---|---|---|---|---|---|---|---|---|---|
| | RMSE | MAE | MAPE | RMSE | MAE | MAPE | RMSE | MAE | MAPE | RMSE | MAE | MAPE | RMSE | MAE | MAPE |
| TimeGPT | 0.030 | 0.025 | 0.128 | 0.097 | 0.063 | 0.399 | 0.162 | 0.123 | 0.645 | **0.101** | 0.079 | 0.583 | 0.207 | 0.169 | 0.607 |
| MLP | 0.039 | 0.032 | 0.130 | 0.051 | 0.040 | 0.190 | **0.059** | 0.048 | **0.198** | 0.103 | 0.082 | **0.378** | **0.146** | **0.114** | 0.399 |
| LSTM | 0.049 | 0.040 | 0.154 | 0.075 | 0.060 | 0.238 | 0.083 | 0.065 | 0.241 | 0.107 | 0.088 | 0.482 | 0.179 | 0.139 | 0.410 |
| LR | **0.017** | **0.015** | **0.067** | **0.032** | **0.025** | **0.129** | 0.067 | **0.047** | 0.318 | 0.099 | **0.073** | 0.471 | 0.195 | 0.132 | 0.351 |
| XGBoost | 0.080 | 0.068 | 0.222 | 0.068 | 0.051 | 0.210 | 0.073 | 0.057 | 0.241 | 0.102 | 0.078 | 0.408 | 0.152 | 0.106 | **0.283** |
| RT | 0.107 | 0.089 | 0.291 | 0.117 | 0.094 | 0.335 | 0.086 | 0.073 | 0.303 | 0.118 | 0.092 | 0.404 | 0.162 | 0.118 | 0.319 |
| PM | 0.042 | 0.032 | 0.156 | 0.099 | 0.075 | 0.414 | 0.165 | 0.133 | 0.710 | 0.108 | 0.086 | 0.648 | 0.171 | 0.142 | 0.604 |
| PatchTST | 0.080 | 0.067 | 0.412 | 0.122 | 0.098 | 0.679 | 0.129 | 0.099 | 0.734 | 0.136 | 0.103 | 0.856 | 0.178 | 0.142 | 0.755 |
| TimeLLM | 0.237 | 0.203 | 0.598 | 0.244 | 0.200 | 1.082 | 0.354 | 0.302 | 1.601 | 0.157 | 0.125 | 0.890 | 0.305 | 0.250 | 1.180 |

***1) Model Performance in Data-rich Cases***

Tables VIII-XI show that in data-rich cases, the evaluation metrics of TimeGPT are generally larger than those of most machine learning models. For example, in Table VIII, the RMSE, MAE, and MAPE of TimeGPT are larger than those of MLP, LSTM, and LR for load forecasting with a 1-hour look-ahead time. This again highlights that TimeGPT is not suitable for load forecasting when extensive historical load data is available.

***2) Model Performance in Data Scarce Cases***

In Tables XII and XV, TimeGPT usually outperforms the benchmarks in cases with scarce historical data, especially for load forecasting with a short look-ahead time. For example, in Table XII with a 1-hour look-ahead time, the RMSE of TimeGPT is reduced by 59.48%, 51.04%, 44.05%, 24.19%, 21.67%, 11.32%, 53.92%, 68.03%, compared to MLP, LSTM, LR, XGBoost, RT, PM, PatchTST, and TimeLLM, respectively. Similarly, the MAE and MAPE of TimeGPT also are smaller than those of benchmarks. This reaffirms the effectiveness of TimeGPT in load forecasting with a short look-ahead time.

However, in Table XV, the evaluation metrics of TimeGPT are not minimal for load forecasting with various look-ahead times. This could be due to significant distribution differences between the load dataset from Arizona State University and the training data. For example, if a machine learning model is trained on images to recognize cats and dogs, its performance will be very limited when transferred to the task of tumor recognition in medical imaging. This is because the data differences between animal images and medical images are too significant. In fact, TimeGPT can be seen as a classic application of transfer learning based on the pre-training and fine-tuning paradigm. TimeGPT first learns the general knowledge from massive and diverse time series data, and then adapts to specific tasks through fine-tuning. Therefore, the performance of TimeGPT in load forecasting is influenced by the differences between datasets in the target and the source domains. If these differences are large, TimeGPT cannot generalize well to load forecasting tasks.

In transfer learning, domain adaptation techniques are commonly used to reduce the distributional differences between source and target domains [43]. However, the domain adaptation techniques are not suitable for TimeGPT and other LTSMs. In LTSMs, the source domain contains massive data, while the load data in the target domain is scarce. This imbalance between the source and target domains can cause domain adaptation techniques to be ineffective, resulting in poor performance of LTSMs in the target domain. One possible solution is to train a specialized LTSM for the energy domain using massive energy data. This can help to maintain the generalization ability of the model while reducing the distributional differences between the source and target domains.

The result in Table XV illustrate that it cannot be guaranteed that TimeGPT is always superior to benchmarks for load forecasting with scarce data. In other words, although TimeGPT performs well on multiple datasets (e.g., load data from China Nongfu Spring Company and University of Texas at Austin), it cannot be ensured to always be the best choice for load forecasting with scarce data.

Overall, TimeGPT has shown strong potential and good performance in load forecasting with scarce historical data on several datasets (e.g., datasets from China Nongfu Spring Company and University of Texas at Austin). However, TimeGPT may not perform as well as the benchmarks on some datasets, possibly due to a significant difference between the current load dataset and the training data.

## V. DISCUSSION

To summarize the performance of TimeGPT in load forecasting, Table XVI compares TimeGPT and benchmark models across different datasets under data scarce cases (i.e., three days of training data). In the first four datasets, TimeGPT outperforms all the benchmarks, particularly for short forecast horizons (e.g., 1 hour). However, TimeGPT does not show an advantage in the fifth dataset, likely due to differences between source and target domains, as explained in detail in Section IV.

In practical applications, operators can use the following strategy to determine whether to use TimeGPT for load forecasting with scarce historical data:

Firstly, the historical data can be divided into a training set and a validation set. Secondly, the training set is used to fine-tune the weights of TimeGPT, while the validation set is used to test the performance of TimeGPT and benchmarks. If TimeGPT outperforms the benchmarks on the validation set (i.e., the evaluation metrics of TimeGPT is minimal on the validation set), then TimeGPT would be the optimal choice for load forecasting with scarce historical data.

TABLE XVI

COMPARISON OF TIMEGPT AND BENCHMARKS ACROSS DATASETS

| Dataset | LAT=1 h | LAT=4 h | LAT=6 h | LAT=12 h | LAT=48 h |
|---|---|---|---|---|---|
| Texas dataset in [36] | √ | √ | √ | √ | √ |
| Nongfu Dataset in [40] | √ | √ | √ | √ | √ |
| Midea Dataset in [40] | √ | × | × | × | × |
| Joho city dataset in [41] | √ | √ | × | × | × |
| Arizona state dataset in [42] | × | × | × | × | × |

LAT: look-head time. √ : Proposed method outperforms all the benchmarks. ×: Proposed method does not outperform all the benchmarks.

# VI. CONCLUSION

This paper explores the potential of large time series models, specifically TimeGPT, in load forecasting with limited historical data. Key findings from simulations and analyses on various datasets are summarized as follows:

While TimeGPT is trained on massive and diverse datasets totaling 100 billion data points, fine-tuning is essential to adapt it effectively for load forecasting tasks.

TimeGPT shows inferior performance to popular machine learning models when abundant historical data is available. In data-scarce scenarios, TimeGPT generally outperforms benchmarks, particularly for short forecast horizons. This highlights the superiority of TimeGPT for load forecasting in scenarios where data is scarce due to the various reasons, such as privacy concerns. However, its performance may be affected by disparities between the training data and the target dataset, emphasizing the need to consider data distribution when applying TimeGPT. In practical applications, operators can divide the historical data into a training set and a validation set, and then use the validation set loss to decide whether TimeGPT is the best choice for a specific dataset.

Although TimeGPT demonstrates strong potential in load forecasting with limited historical data, it relies exclusively on historical load data, which restricts its ability to integrate additional relevant information, such as weather forecasts. To improve forecast accuracy, weather forecasts will be integrated into TimeGPT in future work.

**Funding**

This work is funded by the Swiss Federal Office of Energy (Grant No. SI/502135–01). Also, this work is carried out in the frame of the “UrbanTwin: An urban digital twin for climate action: Assessing policies and solutions for energy, water and infrastructure” project with the financial support of the ETH-Domain Joint Initiative program in the Strategic Area Energy, Climate and Sustainable Environment.